\documentclass[fleqn,10pt]{wlscirep}
\usepackage[utf8]{inputenc}
\usepackage[T1]{fontenc}
\usepackage{graphicx}
\usepackage{placeins}
\usepackage{siunitx}
\usepackage{booktabs}
\usepackage{hyperref}
\usepackage{lineno}

\title{Computed tomography coronary angiogram images, annotations and associated data of normal and diseased arteries}

\author[1]{Ramtin Gharleghi}
\author[2,3]{Dona Adikari}
\author[2,3]{Katy Ellenberger}
\author[4]{Mark Webster}
\author[4]{Chris Ellis}
\author[1]{Arcot Sowmya}
\author[2,3]{Sze-Yuan Ooi}
\author[1]{Susann Beier}

\affil[1]{Faculty of Engineering, University of New South Wales, Sydney, Australia}
\affil[2]{Prince of Wales Clinical School of Medicine, UNSW Sydney, Sydney, Australia}
\affil[3]{Department of Cardiology, Prince of Wales Hospital, Sydney, Australia}
\affil[4]{Auckland City Hospital, Auckland, New Zealand}

\affil[*]{corresponding author(s): Ramtin Gharleghi (r.gharleghi@unsw.edu.au)}

\begin{abstract}
Computed Tomography Coronary Angiography (CTCA) is a non-invasive method to evaluate coronary artery anatomy and disease. CTCA is ideal for geometry reconstruction to create virtual models of coronary arteries. To our knowledge there is no public dataset that includes centrelines and segmentation of the full coronary tree.

We provide anonymized CTCA images, voxel-wise annotations and associated data in the form of  centrelines, calcification scores and meshes of the coronary lumen in 20 normal and 20 diseased cases. Images were obtained along with patient information with informed, written consent as part of Coronary Atlas (https://www.coronaryatlas.org/). Cases were classified as normal (zero calcium score with no signs of stenosis) or diseased (confirmed coronary artery disease). Manual voxel-wise segmentations by three experts were combined using majority voting to generate the final annotations.

Provided data can be used for a variety of research purposes, such as 3D printing patient-specific models, development and validation of segmentation algorithms, education and training of medical personnel and \textit{in-silico} analyses such as testing of medical devices.

\end{abstract}
\begin{document}

\flushbottom
\maketitle

\thispagestyle{empty}

\section*{Background \& Summary}

Coronary artery disease is a leading cause of death worldwide \cite{world2012atlas}, causing a large body of research to focus on the understanding of coronary anatomy and blood flow, disease progression and treatment options \cite{garcia2009computed, goodacre2013systematic, van2018ctca}. With rapid advancements in computation, additive manufacturing and other technologies capable of taking advantage of virtual organ models, computational models of coronary arteries have been increasingly used in research, including the designing and testing of medical devices, as well as for education and training purposes \cite{li2017human}. 

While different modalities can be used to image coronary arteries, only Computed Tomography Coronary Angiography (CTCA) is non-invasive and has sufficient sub-millimetre resolution to allow reconstruction of the small coronary arteries. Therefore it is commonly used and as a result ideal as underlying modality for subsequent image segmentation and virtual coronary artery reconstruction. This commonly required manual refinement after initial automatic threshold due to the small scale, lack of clear contrast with the surrounding tissue and common image artefacts, especially for calcified lesions. Segmentation of the full coronary tree is particularly difficult as even with highest resolution CTCA machines today, distal vessels are only captured via a few image pixels. As a result, despite a wealth of CTCA data available to date, there are extremely few virtual coronary models publicly available and the use of reconstruction workflows on a large-scale patient-specific basis is cost and time intensive. 

Traditional segmentation methods are extremely time consuming \cite{moccia2018blood}, generally requiring semi-automated segmentation closely supervised by a human expert to guide the algorithm and correct errors. Additionally, the segmentations produced are highly sensitive to the individual expert and hence consistent segmentation between different experts is difficult. This has led to no public datasets currently available for use in applications that require accurate patient-specific coronary models. Related datasets are limited, including the `Visible Heart Project', which, however focuses on educational images and videos using magnetic resonance imaging. Although access may be provided to limited CTCA images, these are without annotations or reconstructed models \cite{RN2}. Also available is the 'The Rotterdam dataset' \cite{schaap2009standardized,RN1}, which is primary public dataset, but focused on stenosis detection and stenosis evaluation with sub-voxel accuracy. This dataset may only be used for its stated purpose of stenosis detection and lumen segmentation, and is also no longer publicly available from the challenge website \cite{rotterdamweb}.

To overcome the problems with these traditional segmentation methods, we created high quality segmentations of the coronary arteries, to serve as both a benchmark dataset for newly developed segmentation methods and pre-existing segmentation for further processing, for example investigating differences in helicity between stented idealized and patient-specific vessels \cite{shen2021secondary}. This was part of the `Automated Segmentation of Coronary Arteries' (ASOCA) Challenge \cite{cmig,ramtin_gharleghi_2020_3819799} we facilitated during the Medical Image Computing and Computer Assisted Intervention (MICCAI) 2020 conference to focus on the development of automated segmentation algorithms using this data, providing a convenient system for submission of results and automated evaluation and ranking.

The coronary artery CTCA images were available to us through the Coronary Atlas \cite{medrano2014construction}, an ongoing collection of CTCA images and associated clinical and demographics data used to investigate differences in coronary anatomy \cite{medrano2017study} and haemodynamic behaviour between patients \cite{RN9,beier2016impact,beier2017vascular}. A set of 40 patient-specific coronary artery tree data is provided here, including anonymized CTCA images in .nrdd format, combined high-quality manual voxel annotations derived from 3 experts, and other associated data such as centrelines, smoothed  meshes in .stl format and calcification scores. These served as the training dataset for the ASOCA challenge. Our dataset is the only public dataset of annotations and associated data of the full coronary tree in 20 normal and 20 disease cases. Additionally, a separate set of 20 CTCA images (the test set images for the ASOCA challenge) is provided primarily to facilitate participating in the challenge. In order to not compromise the integrity of the challenge, no other information is provided with these images. Researchers can participate on the challenge website (\url{asoca.grand-challenge.org}), using the training data to develop segmentation algorithms and submit results to the challenge website for automatic evaluation and scoring.

In summary, the current dataset has several advantages over previously available coronary artery datasets. While our dataset is based solely on CTCA and can not provide sub-voxel segmentation and stenosis identification as accurate as the Rotterdam dataset, we do however provide high quality segmentation of all coronary vessels visible in CTCA. In contrast our dataset is available too all researchers including commercial projects. Further, our inclusion of all arteries larger than 1 mm  rather than selected vessel segments allows for expanded applications such as more complex simulations, and more comprehensive training and educational applications. The balanced set of normal and diseased patients ensures effects of disease can be independently studied, as well as ensuring that newly developed segmentation algorithms can robustly handle cases with disease. The dataset is sufficiently large and balanced for training machine learning models. Device manufacturers and researchers with an interest in cardiovascular modelling, prediction and treatment of coronary artery disease can analyse this data directly or combine it with other available datasets. The smooth surface meshes  and centrelines can be directly used for  computational modelling \cite{beier2016impact}, directly 3D printed for experiments \cite{gharleghi20213d,wang2016dual,beier2016dynamically, RN7}, assist in developing and testing medical devices such as stents \cite{antoine2016stentable,zhong2018application,RN5}, and can be used for Virtual Reality applications for education and training \cite{RN6,dugas2016advanced,silva2018emerging}. Moreover, our dataset allows for the development and benchmarking of new segmentation algorithms aiming to efficiently annotate the coronary arteries automatically as per ASOCA challenge \cite{ramtin_gharleghi_2021_4460628}.

\section*{Methods}
\subsection*{Patient Cohort}
Forty patients were randomly selected from a retrospective dataset based on the calcification, stenosis and image quality reported by the cardiologist. Images must have acceptable quality as described by the cardiologist. The dataset was divided into twenty normal patients with no evidence of stenosis and non-obstructive disease, and twenty patients with evidence of calcium scores higher than 0 and obstructive disease. The calcification score in the diseased group ranged between 1 and 986 with a mean of 254. Obstructions in the diseased group ranged from 30\% to 70\% stenosis. Patients were included during routine procedures after written and informed consent and approval from University of New South Wales Human Research Ethics Committee (Ref. 022961).

\subsection*{Imaging}
The CTCA imaging was undertaken using a multi-detector CT scanner (GE Lightspeed 64 multi-slice scanner, USA) using retrospective ECG gating. A contrast medium (Omnipaque 350) was used for imaging and the patient heart rate was controlled around 60bpm by administration of beta blockers. The end diastolic time step was saved for analysis and the images exported as DICOM files. Images were converted to Nearly Raw Raster Data (NRRD) format during the anonymization process and the intensity rescaled to Hounsfield units based on the appropriate DICOM tags. 

\subsection*{Annotation}
\begin{figure}
	\centering
	\includegraphics[width=0.8\linewidth]{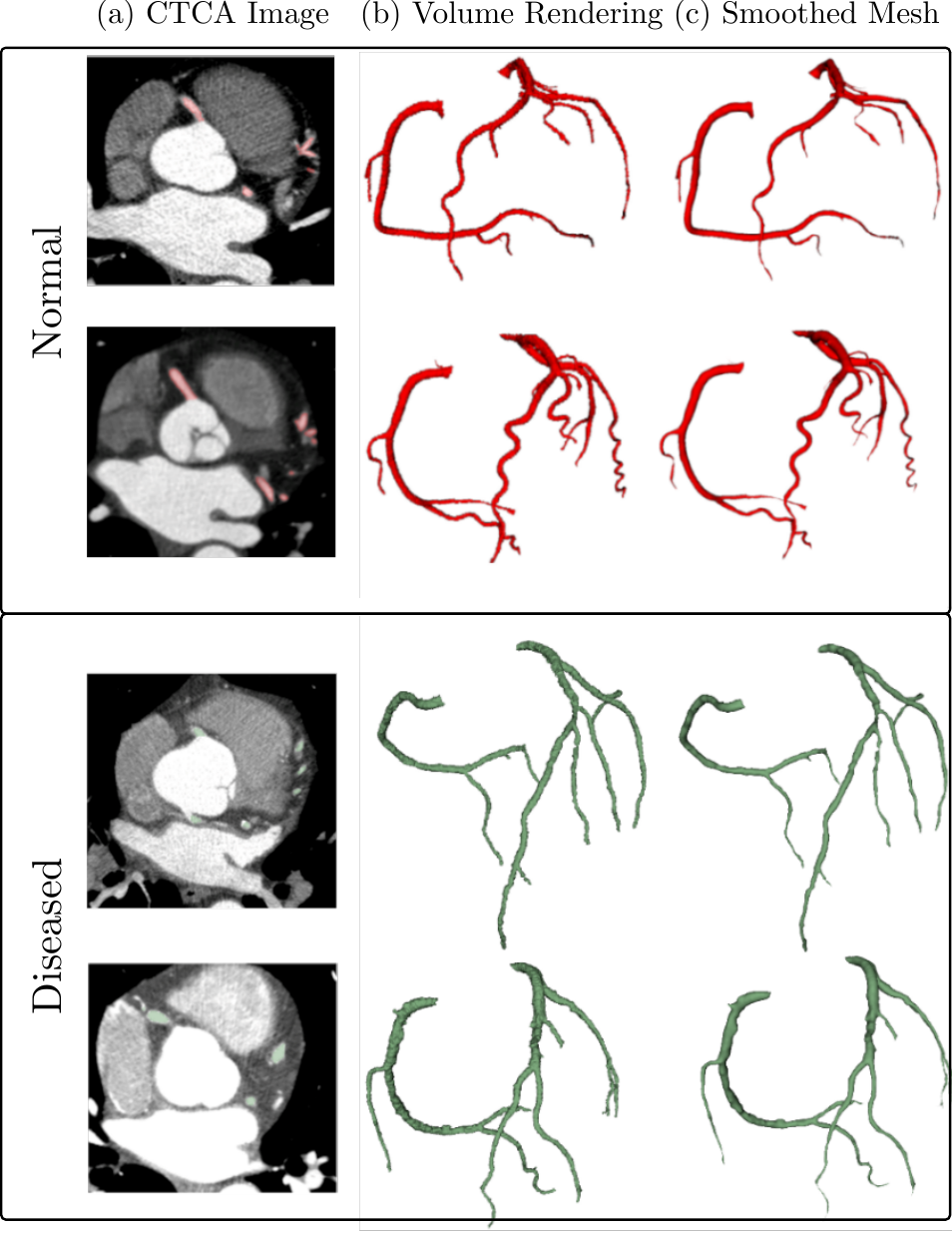}
	\caption{Samples of annotated data showing (a) an annotated slice of the CTCA images, (b) volumetric rendering of the labelled voxels, and (c) smooth surface mesh (left to right) generated from the normal (top two rows) and diseased (bottom two rows) coronary artery image annotations.}
	\label{fig:image1}
\end{figure}

\begin{figure}
	\centering
	\includegraphics[width=1\linewidth]{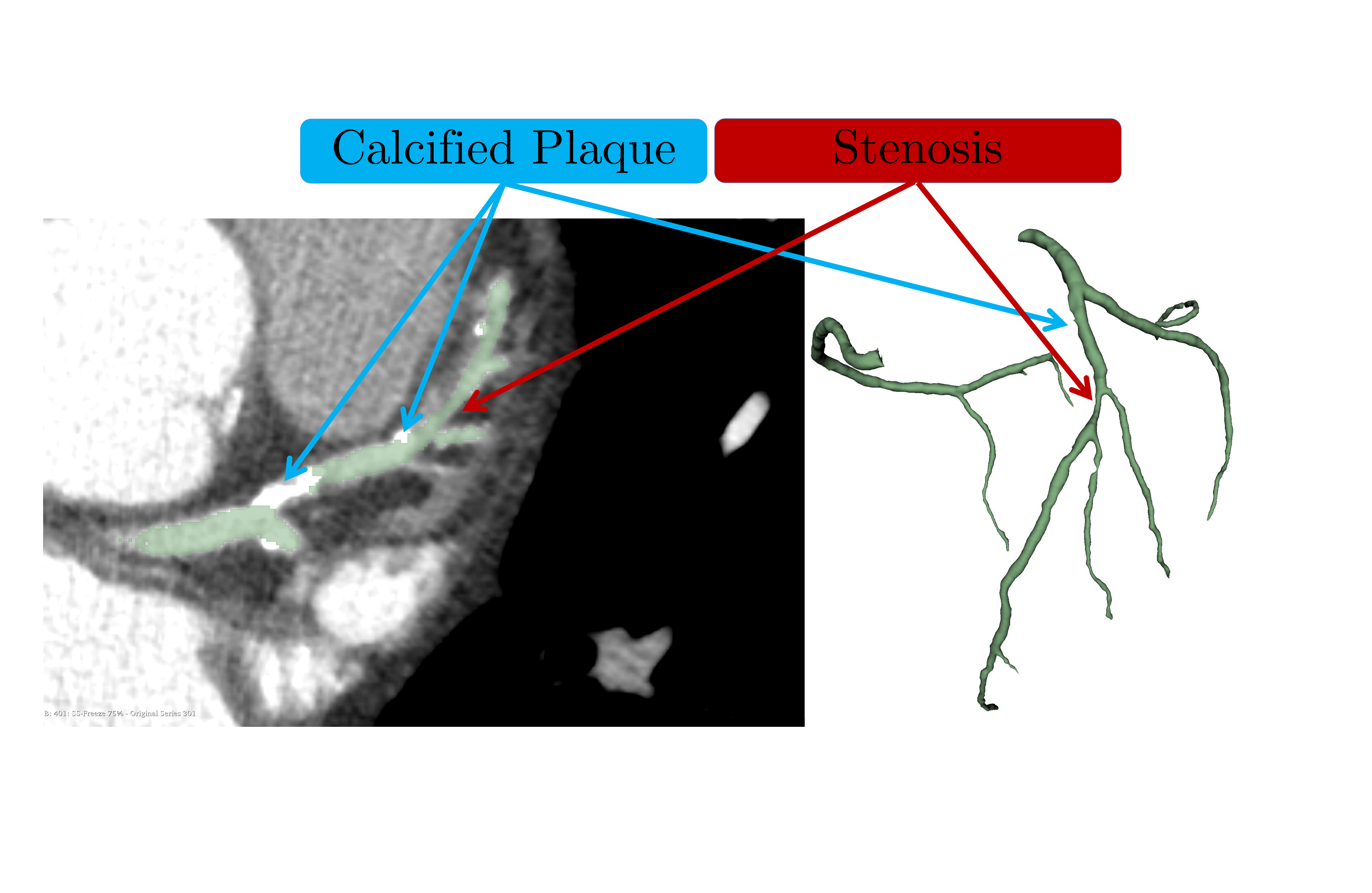}
	\caption{Calcified and non-calcified plaques present in the dataset.}
	\label{fig:image2}
\end{figure}

The open-source software 3D Slicer (version 4.3) \cite{RN10} was used to manually annotate the coronary arteries images. The annotation process was performed independently by three annotators, who were instructed to segment the left and right coronary trees starting at the aortic root. Thresholding at a cut-off chosen by the expert was used to generate an initial rough segmentation of the vessel, followed by manually correcting the vessel contours in each slice. All coronary vessels with a diameter larger 1 mm, representing 1-2 voxels, were included in the segmentation. In segments showing significant imaging artefacts affecting the vessel that would make further segmentation unfeasible, the rest of the vessel was ignored. A sample of the annotated CTCA images and the resulting 3D reconstructions are shown in Figure \ref{fig:image1}.  Figure \ref{fig:image2} shows a diseased case with calcified plaque and stenosis. 

\subsection*{Post Processing}
\begin{figure}
	\centering
	\includegraphics[width=0.5\linewidth]{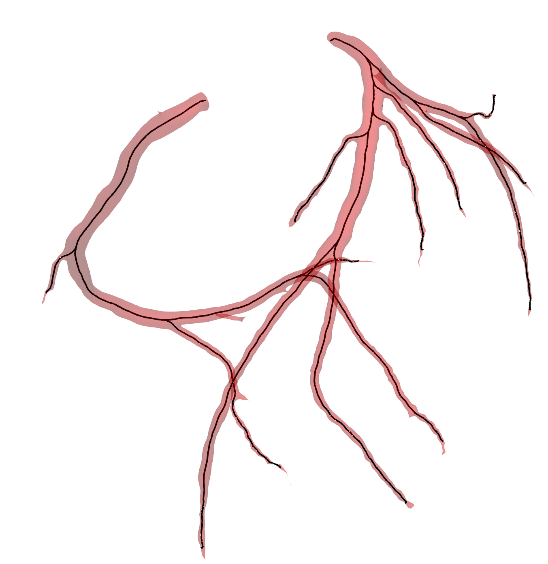}
	\caption{Sample coronary tree surface and centreline.}
	\label{fig:image3}
\end{figure}

The annotations are combined to produce a final segmentation of the arteries by majority vote among the three annotations, i.e. including regions where at least two of the annotators agreed. Small vessels (\textless1 mm, i.e. 1-2 voxels) were removed if they were mistakenly included. The segmentations are available as voxel-wise annotations, as well as smoothed surface meshes.    Surface meshes were produced from the annotations using the Flying Edges algorithm \cite{RN11}. It should be noted that with voxel-wise labelling as used in this dataset rather than a tubular parametrization, further smoothing would be necessary to recover a smooth vessel shape. The annotations were smoothed using Taubin’s algorithm \cite{RN12}, implemented in the open-source Vascular Modelling Tool Kit (VMTK,  https://www.vmtk.org ) \cite{RN8,izzo2018vascular}, with a passband of 0.03 and 30 iterations before being exported as an STL file. Taubin's smoothing method is commonly used when processing vessel segmentations \cite{shum2011framework} and is expected to preserve topology and volume of the vessels \cite{antiga2008image}.These settings correspond to the smoothing used in the Coronary Atlas to calculate shape parameters. The raw annotations provided can be used to produce surface meshes with different smoothing settings if needed.
Vessel centrelines were extracted manually by marking the inlet and outlet points on the mesh for automated centreline calculation in VMTK, as shown in Figure \ref{fig:image3}. 

\subsection*{ASOCA Test Data Set}
An additional 10 normal and 10 diseased CTCA cases, separate to the 20 normal and 20 diseased used for the training data, were selected based on the same criteria to serve as the test set for the ASOCA challenge. These cases will be distributed alongside the annotated dataset to facilitate further participation in the challenge. Ground truth annotation and other associated data for these cases is not publicly available.

\section*{Data Records}
Obtaining access to the dataset is described on the challenge website (\url{https://asoca.grand-challenge.org/access/}). Patients are labelled sequentially from 1 to 20, with normal and diseased patients labelled separately (i.e. Normal\_1…Normal\_20 represent the normal patients and Diseased\_1…Diseased\_20 represent diseased patients). CTCA scans are provided as Nearly Raw Raster Data (NRRD) file labelled sequentially based on patient name (Normal\_1.nrrd, Normal\_2.nrrd…). This naming convention is used for the rest of the data folders. The annotations folder contains the final annotation for each patient. This represents the voxel-wise annotations, with the background voxels assigned a value of 0 and the foreground (vessel lumen) assigned a value of 1. Both the CTCA images and annotations have anisotropic resolution, a common characteristic of most CT machines, with the z-axis resolution of 0.625mm and the in-plane resolution ranging from 0.3mm to 0.4mm depending on the patient. The SurfaceMeshes directory contains smooth surface meshes generated from the voxel annotations. These meshes are provided in STL format, with an average of 37,000 vertices to capture the arterial geometry.

The centrelines folder contains centrelines of the coronary arteries for each patient, provided in VTK Poly Data (VTP) format that allows for efficient storage of centreline data. Figure 3 shows a sample of the extracted centreline and underlying surface mesh. 
The spreadsheet DiseaseReports.xlsx reports calcium score and stenoses levels for each patient.

\section*{Technical Validation}
Dice Similarity Coefficient (DSC) \cite{bertels2019optimizing} is frequently used to measure the degree of overlap between annotations. DSC is defined as in eq. \ref{dsc}  for two sets of voxels A and B. Similarly, Hausdorff Distance (HD) as shown in eq. \ref{hd} measures the distance of corresponding points between annotations. In practice commonly the 95th percentile HD is used rather than the maximum in order to reduce sensitivity to outliers \cite{taha2015metrics}.
\begin{equation}
	\label{dsc}
	\text{DSC}=\frac{2|A \cap B}{|A|+|B|}
\end{equation} 
\begin{equation}
	\label{hd}
   \text{HD} = \max(\max_{x\in A} \min_{y\in B} d(x,y), \max_{y\in A} \min_{x\in B} d(x,y))
\end{equation}
We used DSC (Table 1) and 95\textsuperscript{th} percentile HD (Table 2) to compare variability between annotators compared to the final ground truth generated for each case. The average Dice Score among the three annotators was 85.6\%$\pm$7.7\% (mean$\pm$standard deviation) and an average HD of 5.92$\pm$7.3 mm (mean$\pm$standard deviation). The concordance between annotators was higher for normal cases compared to diseased (87.4\% vs 83.9\%, p=0.01 using Welch's t test \cite{derrick2016welch}), due presence of stenosis and calcified plaques complicating the annotation of diseased images. Hausdorff Distance showed similar results (4.45 mm in normal cases vs 7.38 mm in diseased, p=0.028). A Dice Score of 1 (indicating perfect agreement) is difficult to achieve, as this dataset attempts to segment the full coronary artery tree including small arteries near the limit of CTCA imaging resolution. This Dice Score and Hausdorff Distance indicates high agreement between the annotators and is unlikely to adversely affect usage of this dataset. Table 3 shows the Hausdorff Distance between centre of the voxel labels and the smoothed mesh.

\begin{table}
	\centering
	\sisetup{table-parse-only}
	\caption{Annotator Dice Similarity Coefficient for each patient.}
	\resizebox{\textwidth}{!}{%
	\rowcolors{2}{gray!25}{white}
	\begin{tabular}{cSSS|cSSS}
		\multicolumn{4}{c}{\textbf{Normal}}                        & \multicolumn{4}{c}{\textbf{Diseased}}                      \\
		Patient & {Annotator 1 (\%)}& {Annotator 2 (\%)}& {Annotator 3 (\%)}& Patient & {Annotator 1 (\%)}& {Annotator 2 (\%)}& {Annotator 3 (\%)}\\\hline
		\#1       & 95.1     & 91.8     & 92.3     & \#1       & 84.2     & 82.9     & 86.0     \\
		\#2       & 79.3     & 82.4     & 93.0     & \#2       & 84.7     & 81.1     & 86.6     \\
		\#3       & 96.7     & 85.9     & 77.3     & \#3       & 71.8     & 83.6     & 86.0     \\
		\#4       & 96.7     & 75.6     & 81.3     & \#4       & 92.7     & 89.8     & 67.7     \\
		\#5       & 86.1     & 90.4     & 93.4     & \#5       & 96.3     & 87.7     & 83.1     \\
		\#6       & 91.7     & 81.2     & 97.4     & \#6       & 83.9     & 83.8     & 87.2     \\
		\#7       & 91.6     & 86.4     & 93.6     & \#7       & 84.9     & 76.3     & 91.3     \\
		\#8       & 87.8     & 82.4     & 90.5     & \#8       & 86.4     & 79.4     & 83.4     \\
		\#9       & 97.7     & 73.5     & 84.9     & \#9       & 90.3     & 82.0     & 84.0     \\
		\#10      & 95.7     & 89.8     & 95.0     & \#10      & 82.2     & 79.1     & 84.0     \\
		\#11      & 88.3     & 93.5     & 86.1     & \#11      & 84.2     & 80.5     & 85.3     \\
		\#12      & 92.4     & 78.9     & 87.6     & \#12      & 80.4     & 82.2     & 85.9     \\
		\#13      & 98.1     & 92.8     & 70.4     & \#13      & 83.1     & 87.6     & 79.6     \\
		\#14      & 96.2     & 90.2     & 57.2     & \#14      & 88.3     & 89.7     & 81.4     \\
		\#15      & 98.0     & 77.9     & 67.3     & \#15      & 76.4     & 85.7     & 85.1     \\
		\#16      & 98.4     & 72.8     & 93.8     & \#16      & 79.4     & 76.9     & 89.4     \\
		\#17      & 97.0     & 78.4     & 85.3     & \#17      & 90.3     & 88.7     & 71.8     \\
		\#18      & 91.9     & 92.8     & 60.2     & \#18      & 78.1     & 90.6     & 88.1     \\
		\#19      & 87.2     & 86.2     & 92.7     & \#19      & 80.8     & 85.0     & 85.4     \\
		\#20      & 90.5     & 92.2     & 97.4     & \#20      & 82.0     & 87.3     & 83.3    
	\end{tabular}
	}
\end{table}
\begin{table}
	\centering
	\sisetup{table-parse-only}
	\caption{Annotator 95\textsuperscript{th} percentile Hausdorff Distance for each patient.}
	\resizebox{\textwidth}{!}{
	  \rowcolors{2}{gray!25}{white}
	\begin{tabular}{cSSS|cSSS}
		\multicolumn{4}{c}{\textbf{Normal}}                        & \multicolumn{4}{c}{\textbf{Diseased}}                      \\
		Patient & {Annotator 1 [mm]}& {Annotator 2 [mm]}& {Annotator 3 [mm]}& Patient & {Annotator 1 [mm]}& {Annotator 2 [mm]}& {Annotator 3 [mm]}\\\hline
\#1  & 0.42  & 4.07  & 0.42  & \#1 & 9.2   & 5.12  & 2.3   \\
\#2  & 1.1   & 4.86  & 0.62  & \#2 & 4.0   & 14.36 & 0.44  \\
\#3  & 0.0   & 3.56  & 5.58  & \#3 & 11.91 & 3.61  & 0.62  \\
\#4  & 0.0   & 5.74  & 0.62  & \#4 & 0.62  & 10.29 & 7.29  \\
\#5  & 0.72  & 3.42  & 8.16  & \#5 & 0.45  & 11.31 & 5.55  \\
\#6  & 2.17  & 7.82  & 0.0   & \#6 & 0.62  & 10.08 & 1.4   \\
\#7  & 6.92  & 1.31  & 0.4   & \#7 & 0.56  & 12.12 & 0.56  \\
\#8  & 0.57  & 1.49  & 4.78  & \#8 & 37.46 & 15.92 & 14.56 \\
\#9  & 0.0   & 21.78 & 0.97  & \#9 & 7.76  & 5.56  & 2.82  \\
\#10 & 0.44  & 10.79 & 8.11  & \#10 & 0.73  & 3.32  & 11.66 \\
\#11 & 2.92  & 1.4   & 1.78  & \#11 & 0.7   & 12.11 & 1.68  \\
\#12 & 0.73  & 12.61 & 0.52  & \#12 & 2.34  & 6.01  & 0.7   \\
\#13 & 0.0   & 0.33  & 2.35  & \#13 & 41.0  & 3.39  & 1.53  \\
\#14 & 0.37  & 2.4   & 13.87 & \#14 & 14.37 & 8.47  & 1.21  \\
\#15 & 0.0   & 9.98  & 10.51 & \#15 & 1.8   & 6.74  & 0.9   \\
\#16 & 0.0   & 16.87 & 0.35  & \#16 & 29.7  & 15.0  & 0.62  \\
\#17 & 0.36  & 6.14  & 21.86 & \#17 & 0.39  & 5.3   & 15.92 \\
\#18 & 10.86 & 3.57  & 19.62 & \#18 & 21.26 & 2.6   & 0.65  \\
\#19 & 3.15  & 13.76 & 0.38  & \#19 & 5.13  & 13.0  & 3.16  \\
\#20 & 0.42  & 3.29  & 0.0   & \#20 & 4.87  & 6.87  & 3.64 
	\end{tabular}
	}
\end{table}

\begin{table}
\centering
\caption{95$^{th}$ percentile Hausdorff Distance between smoothed meshes and voxel labelmap}
\resizebox{\textwidth}{!}{%
  \rowcolors{2}{gray!25}{white}
\begin{tabular}{cllllllllllllllllllll}
Patient \#&\#1	&\#2&\#3&\#4&\#5&\#6&\#7&\#8&\#9&\#10&\#11&\#12&\#13&\#14&\#15&\#16&\#17&\#18&\#19 &	\#20 \\[15pt]\hline
Normal [mm] & 1.39      & 1.13      & 1.40      & 1.05      & 1.16      & 1.45      & 1.35      & 1.40      & 1.24      & 1.49       & 1.28       & 1.24       & 1.31       & 1.24       & 1.40       & 1.38       & 1.22       & 1.25       & 1.13       & 1.20       \\[15pt]
Diseased [mm] & 1.10      & 1.03      & 1.15      & 1.40      & 1.36      & 1.14      & 1.19      & 1.06      & 1.30      & 1.08       & 1.17       & 1.17       & 1.29       & 1.28       & 1.05       & 1.43       & 1.05       & 1.51       & 1.31       & 1.34 \\[15pt]
\end{tabular}%
}
\end{table}
\section*{Usage Notes}
These recommendations focus on free, open-source software, however as the dataset is provided commonly used formats commercially available software suites will can also be utilised.
CTCA and ground-truth data is provided in NRRD format, compatible with all common medical imaging software such as 3D Slicer\cite{RN10} and ITK-SNAP \cite{yushkevich2006user}. 3D Slicer is the recommended software for working with this data, providing tools for common editing operations and various add-ons for specialised tasks. The centrelines are saved in VTK Poly Data (VTP) format, expected to be used with the Visualization Toolkit (VTK) \cite{VTK4} and the Vascular Modelling Toolkit \cite{RN8,izzo2018vascular}. VMTK is also available as a 3D Slicer add-on. Surface meshes are provided in Standard Tessellation Language (STL), compatible with most mesh software. Both 3D Slicer and VMTK allow editing and processing STL meshes, including addition of flow extensions and generation of volume meshes for computational fluid dynamics simulations. Specific mesh editing software such as Meshlab \cite{meshlab} can be used for more complex tasks. The dataset can be also be used to develop new segmentation algorithms and evaluate the performance on the standardised ASOCA challenge. Submission instructions are available on the challenge website ( \url{https://asoca.grand-challenge.org/SubmittingResults/}). 

The dataset can be used for unrestricted research purposes. Researchers should follow the process described on the challenge website (\url{https://asoca.grand-challenge.org/access/}).

\section*{Code availability}
The code for creation of this dataset, usage examples and evaluation code used in the challenge is available on GitHub (\url{https://github.com/Ramtingh/ASOCADataDescription}.
Figure 1, 2 and 3 were created with data included in the dataset. A copy of the raw data used  is included in the repository under the corresponding folder to maker recreating these figures easier. 3D Slicer (version 4.3) was used in the preparation of the dataset and Figures 1 and 2. Vascular Modelling Tool Kit (version 1.4) was used to calculate centerlines and generate Figure 3. 

\FloatBarrier
\bibliography{refs}

\begin{thebibliography}{10}
\urlstyle{rm}
\expandafter\ifx\csname url\endcsname\relax
  \def\url#1{\texttt{#1}}\fi
\expandafter\ifx\csname urlprefix\endcsname\relax\def\urlprefix{URL }\fi
\expandafter\ifx\csname doiprefix\endcsname\relax\def\doiprefix{DOI: }\fi
\providecommand{\bibinfo}[2]{#2}
\providecommand{\eprint}[2][]{\url{#2}}

\bibitem{world2012atlas}
\bibinfo{author}{{World Health Organization}}.
\newblock \bibinfo{journal}{\bibinfo{title}{The atlas of heart disease and
  stroke}}.
\newblock {\emph{\JournalTitle{World Health Organization}}}
  (\bibinfo{year}{2012}).

\bibitem{garcia2009computed}
\bibinfo{author}{Garc{\'\i}a-Garc{\'\i}a, H.~M.} \emph{et~al.}
\newblock \bibinfo{journal}{\bibinfo{title}{Computed tomography in total
  coronary occlusions (ctto registry): radiation exposure and predictors of
  successful percutaneous intervention.}}
\newblock {\emph{\JournalTitle{EuroIntervention: journal of EuroPCR in
  collaboration with the Working Group on Interventional Cardiology of the
  European Society of Cardiology}}} \textbf{\bibinfo{volume}{4}},
  \bibinfo{pages}{607--616} (\bibinfo{year}{2009}).

\bibitem{goodacre2013systematic}
\bibinfo{author}{Goodacre, S.} \emph{et~al.}
\newblock \bibinfo{journal}{\bibinfo{title}{Systematic review, meta-analysis
  and economic modelling of diagnostic strategies for suspected acute coronary
  syndrome}}.
\newblock {\emph{\JournalTitle{Health Technol Assess}}}
  \textbf{\bibinfo{volume}{17}}, \bibinfo{pages}{1--188}
  (\bibinfo{year}{2013}).

\bibitem{van2018ctca}
\bibinfo{author}{van~den Boogert, T.} \emph{et~al.}
\newblock \bibinfo{journal}{\bibinfo{title}{Ctca for detection of significant
  coronary artery disease in routine tavi work-up}}.
\newblock {\emph{\JournalTitle{Netherlands Heart Journal}}}
  \textbf{\bibinfo{volume}{26}}, \bibinfo{pages}{591--599}
  (\bibinfo{year}{2018}).

\bibitem{li2017human}
\bibinfo{author}{Li, Q.} \emph{et~al.}
\newblock \bibinfo{journal}{\bibinfo{title}{An human-computer interactive
  augmented reality system for coronary artery diagnosis planning and
  training}}.
\newblock {\emph{\JournalTitle{Journal of medical systems}}}
  \textbf{\bibinfo{volume}{41}}, \bibinfo{pages}{1--11} (\bibinfo{year}{2017}).

\bibitem{moccia2018blood}
\bibinfo{author}{Moccia, S.}, \bibinfo{author}{De~Momi, E.},
  \bibinfo{author}{El~Hadji, S.} \& \bibinfo{author}{Mattos, L.~S.}
\newblock \bibinfo{journal}{\bibinfo{title}{Blood vessel segmentation
  algorithms—review of methods, datasets and evaluation metrics}}.
\newblock {\emph{\JournalTitle{Computer methods and programs in biomedicine}}}
  \textbf{\bibinfo{volume}{158}}, \bibinfo{pages}{71--91}
  (\bibinfo{year}{2018}).

\bibitem{RN2}
\bibinfo{author}{Iaizzo, P.~A.}
\newblock \bibinfo{journal}{\bibinfo{title}{The visible heart{\textregistered}
  project and free-access website ‘atlas of human cardiac anatomy’}}.
\newblock {\emph{\JournalTitle{EP Europace}}} \textbf{\bibinfo{volume}{18}},
  \bibinfo{pages}{iv163--iv172} (\bibinfo{year}{2016}).

\bibitem{schaap2009standardized}
\bibinfo{author}{Schaap, M.} \emph{et~al.}
\newblock \bibinfo{journal}{\bibinfo{title}{Standardized evaluation methodology
  and reference database for evaluating coronary artery centerline extraction
  algorithms}}.
\newblock {\emph{\JournalTitle{Medical image analysis}}}
  \textbf{\bibinfo{volume}{13}}, \bibinfo{pages}{701--714}
  (\bibinfo{year}{2009}).

\bibitem{RN1}
\bibinfo{author}{Kiri{\c{s}}li, H.} \emph{et~al.}
\newblock \bibinfo{journal}{\bibinfo{title}{Standardized evaluation framework
  for evaluating coronary artery stenosis detection, stenosis quantification
  and lumen segmentation algorithms in computed tomography angiography}}.
\newblock {\emph{\JournalTitle{Medical image analysis}}}
  \textbf{\bibinfo{volume}{17}}, \bibinfo{pages}{859--876}
  (\bibinfo{year}{2013}).

\bibitem{rotterdamweb}
\bibinfo{title}{Rotterdam coronary artery algorithm evaluation framework}.
\newblock \bibinfo{howpublished}{\url{https://coronary.bigr.nl/}}.

\bibitem{shen2021secondary}
\bibinfo{author}{Shen, C.} \emph{et~al.}
\newblock \bibinfo{journal}{\bibinfo{title}{Secondary flow in
  bifurcations--important effects of curvature, bifurcation angle and stents}}.
\newblock {\emph{\JournalTitle{Journal of Biomechanics}}}
  \textbf{\bibinfo{volume}{129}}, \bibinfo{pages}{110755}
  (\bibinfo{year}{2021}).

\bibitem{cmig}
\bibinfo{author}{Gharleghi, R.} \emph{et~al.}
\newblock \bibinfo{journal}{\bibinfo{title}{Automated segmentation of normal
  and diseased coronary arteries - the asoca challenge}}.
\newblock {\emph{\JournalTitle{Computerized Medical Imaging and Graphics}}}
  (\bibinfo{year}{2022}).

\bibitem{ramtin_gharleghi_2020_3819799}
\bibinfo{author}{Gharleghi, R.}, \bibinfo{author}{Samarasinghe, G.},
  \bibinfo{author}{Sowmya, P.~A.} \& \bibinfo{author}{Beier, S.}
\newblock \bibinfo{title}{Automated segmentation of coronary arteries},
  \url{10.5281/zenodo.3819799} (\bibinfo{year}{2020}).

\bibitem{medrano2014construction}
\bibinfo{author}{Medrano-Gracia, P.} \emph{et~al.}
\newblock \bibinfo{title}{Construction of a coronary artery atlas from ct
  angiography}.
\newblock In \emph{\bibinfo{booktitle}{International Conference on Medical
  Image Computing and Computer-Assisted Intervention}},
  \bibinfo{pages}{513--520} (\bibinfo{organization}{Springer},
  \bibinfo{year}{2014}).

\bibitem{medrano2017study}
\bibinfo{author}{Medrano-Gracia, P.} \emph{et~al.}
\newblock \bibinfo{journal}{\bibinfo{title}{A study of coronary bifurcation
  shape in a normal population}}.
\newblock {\emph{\JournalTitle{Journal of cardiovascular translational
  research}}} \textbf{\bibinfo{volume}{10}}, \bibinfo{pages}{82--90}
  (\bibinfo{year}{2017}).

\bibitem{RN9}
\bibinfo{author}{Medrano-Gracia, P.} \emph{et~al.}
\newblock \bibinfo{journal}{\bibinfo{title}{A computational atlas of normal
  coronary artery anatomy}}.
\newblock {\emph{\JournalTitle{EuroIntervention: journal of EuroPCR in
  collaboration with the Working Group on Interventional Cardiology of the
  European Society of Cardiology}}} \textbf{\bibinfo{volume}{12}},
  \bibinfo{pages}{845--854} (\bibinfo{year}{2016}).

\bibitem{beier2016impact}
\bibinfo{author}{Beier, S.} \emph{et~al.}
\newblock \bibinfo{journal}{\bibinfo{title}{Impact of bifurcation angle and
  other anatomical characteristics on blood flow--a computational study of
  non-stented and stented coronary arteries}}.
\newblock {\emph{\JournalTitle{Journal of biomechanics}}}
  \textbf{\bibinfo{volume}{49}}, \bibinfo{pages}{1570--1582}
  (\bibinfo{year}{2016}).

\bibitem{beier2017vascular}
\bibinfo{author}{Beier, S.} \emph{et~al.}
\newblock \bibinfo{title}{Vascular hemodynamics with computational modeling and
  experimental studies}.
\newblock In \emph{\bibinfo{booktitle}{Computing and Visualization for
  Intravascular Imaging and Computer-Assisted Stenting}},
  \bibinfo{pages}{227--251} (\bibinfo{publisher}{Elsevier},
  \bibinfo{year}{2017}).

\bibitem{gharleghi20213d}
\bibinfo{author}{Gharleghi, R.} \emph{et~al.}
\newblock \bibinfo{journal}{\bibinfo{title}{3d printing for cardiovascular
  applications: From end-to-end processes to emerging developments}}.
\newblock {\emph{\JournalTitle{Annals of Biomedical Engineering}}}
  \bibinfo{pages}{1--21} (\bibinfo{year}{2021}).

\bibitem{wang2016dual}
\bibinfo{author}{Wang, K.} \emph{et~al.}
\newblock \bibinfo{journal}{\bibinfo{title}{Dual-material 3d printed
  metamaterials with tunable mechanical properties for patient-specific
  tissue-mimicking phantoms}}.
\newblock {\emph{\JournalTitle{Additive Manufacturing}}}
  \textbf{\bibinfo{volume}{12}}, \bibinfo{pages}{31--37}
  (\bibinfo{year}{2016}).

\bibitem{beier2016dynamically}
\bibinfo{author}{Beier, S.} \emph{et~al.}
\newblock \bibinfo{journal}{\bibinfo{title}{Dynamically scaled phantom phase
  contrast mri compared to true-scale computational modeling of coronary artery
  flow}}.
\newblock {\emph{\JournalTitle{Journal of Magnetic Resonance Imaging}}}
  \textbf{\bibinfo{volume}{44}}, \bibinfo{pages}{983--992}
  (\bibinfo{year}{2016}).

\bibitem{RN7}
\bibinfo{author}{Yoo, S.-J.}, \bibinfo{author}{Spray, T.},
  \bibinfo{author}{Austin~III, E.~H.}, \bibinfo{author}{Yun, T.-J.} \&
  \bibinfo{author}{van Arsdell, G.~S.}
\newblock \bibinfo{journal}{\bibinfo{title}{Hands-on surgical training of
  congenital heart surgery using 3-dimensional print models}}.
\newblock {\emph{\JournalTitle{The Journal of thoracic and cardiovascular
  surgery}}} \textbf{\bibinfo{volume}{153}}, \bibinfo{pages}{1530--1540}
  (\bibinfo{year}{2017}).

\bibitem{antoine2016stentable}
\bibinfo{author}{Antoine, E.~E.}, \bibinfo{author}{Cornat, F.~P.} \&
  \bibinfo{author}{Barakat, A.~I.}
\newblock \bibinfo{journal}{\bibinfo{title}{The stentable in vitro artery: an
  instrumented platform for endovascular device development and optimization}}.
\newblock {\emph{\JournalTitle{Journal of The Royal Society Interface}}}
  \textbf{\bibinfo{volume}{13}}, \bibinfo{pages}{20160834}
  (\bibinfo{year}{2016}).

\bibitem{zhong2018application}
\bibinfo{author}{Zhong, L.} \emph{et~al.}
\newblock \bibinfo{journal}{\bibinfo{title}{Application of patient-specific
  computational fluid dynamics in coronary and intra-cardiac flow simulations:
  Challenges and opportunities}}.
\newblock {\emph{\JournalTitle{Frontiers in physiology}}}
  \textbf{\bibinfo{volume}{9}}, \bibinfo{pages}{742} (\bibinfo{year}{2018}).

\bibitem{RN5}
\bibinfo{author}{Sun, Z.} \& \bibinfo{author}{Jansen, S.}
\newblock \bibinfo{journal}{\bibinfo{title}{Personalized 3d printed coronary
  models in coronary stenting}}.
\newblock {\emph{\JournalTitle{Quantitative Imaging in Medicine and Surgery}}}
  \textbf{\bibinfo{volume}{9}}, \bibinfo{pages}{1356} (\bibinfo{year}{2019}).

\bibitem{RN6}
\bibinfo{author}{Reinhard~Friedl, M.}
\newblock \bibinfo{title}{Virtual reality and 3d visualizations in heart
  surgery education}.
\newblock In \emph{\bibinfo{booktitle}{The Heart surgery forum}},
  \bibinfo{pages}{03054} (\bibinfo{year}{2001}).

\bibitem{dugas2016advanced}
\bibinfo{author}{Dugas, C.~M.} \& \bibinfo{author}{Schussler, J.~M.}
\newblock \bibinfo{journal}{\bibinfo{title}{Advanced technology in
  interventional cardiology: a roadmap for the future of precision coronary
  interventions}}.
\newblock {\emph{\JournalTitle{Trends in cardiovascular medicine}}}
  \textbf{\bibinfo{volume}{26}}, \bibinfo{pages}{466--473}
  (\bibinfo{year}{2016}).

\bibitem{silva2018emerging}
\bibinfo{author}{Silva, J.~N.}, \bibinfo{author}{Southworth, M.},
  \bibinfo{author}{Raptis, C.} \& \bibinfo{author}{Silva, J.}
\newblock \bibinfo{journal}{\bibinfo{title}{Emerging applications of virtual
  reality in cardiovascular medicine}}.
\newblock {\emph{\JournalTitle{JACC: Basic to Translational Science}}}
  \textbf{\bibinfo{volume}{3}}, \bibinfo{pages}{420--430}
  (\bibinfo{year}{2018}).

\bibitem{ramtin_gharleghi_2021_4460628}
\bibinfo{author}{Gharleghi, R.}
\newblock \bibinfo{title}{{Ramtingh/ASOCA\_MICCAI2020\_Evaluation: MICCAI
  Evaluation}}, \url{10.5281/zenodo.4460628} (\bibinfo{year}{2021}).

\bibitem{RN10}
\bibinfo{author}{Fedorov, A.} \emph{et~al.}
\newblock \bibinfo{journal}{\bibinfo{title}{3d slicer as an image computing
  platform for the quantitative imaging network}}.
\newblock {\emph{\JournalTitle{Magnetic resonance imaging}}}
  \textbf{\bibinfo{volume}{30}}, \bibinfo{pages}{1323--1341}
  (\bibinfo{year}{2012}).

\bibitem{RN11}
\bibinfo{author}{Schroeder, W.}, \bibinfo{author}{Maynard, R.} \&
  \bibinfo{author}{Geveci, B.}
\newblock \bibinfo{title}{Flying edges: A high-performance scalable
  isocontouring algorithm}.
\newblock In \emph{\bibinfo{booktitle}{2015 IEEE 5th Symposium on Large Data
  Analysis and Visualization (LDAV)}}, \bibinfo{pages}{33--40}
  (\bibinfo{organization}{IEEE}, \bibinfo{year}{2015}).

\bibitem{RN12}
\bibinfo{author}{Taubin, G.}, \bibinfo{author}{Zhang, T.} \&
  \bibinfo{author}{Golub, G.}
\newblock \bibinfo{title}{Optimal surface smoothing as filter design}.
\newblock In \emph{\bibinfo{booktitle}{European Conference on Computer
  Vision}}, \bibinfo{pages}{283--292} (\bibinfo{organization}{Springer},
  \bibinfo{year}{1996}).

\bibitem{RN8}
\bibinfo{author}{Antiga, L.} \& \bibinfo{author}{Steinman, D.~A.}
\newblock \bibinfo{journal}{\bibinfo{title}{Robust and objective decomposition
  and mapping of bifurcating vessels}}.
\newblock {\emph{\JournalTitle{IEEE transactions on medical imaging}}}
  \textbf{\bibinfo{volume}{23}}, \bibinfo{pages}{704--713}
  (\bibinfo{year}{2004}).

\bibitem{izzo2018vascular}
\bibinfo{author}{Izzo, R.}, \bibinfo{author}{Steinman, D.},
  \bibinfo{author}{Manini, S.} \& \bibinfo{author}{Antiga, L.}
\newblock \bibinfo{journal}{\bibinfo{title}{The vascular modeling toolkit: a
  python library for the analysis of tubular structures in medical images}}.
\newblock {\emph{\JournalTitle{Journal of Open Source Software}}}
  \textbf{\bibinfo{volume}{3}}, \bibinfo{pages}{745} (\bibinfo{year}{2018}).

\bibitem{shum2011framework}
\bibinfo{author}{Shum, J.}, \bibinfo{author}{Xu, A.},
  \bibinfo{author}{Chatnuntawech, I.} \& \bibinfo{author}{Finol, E.~A.}
\newblock \bibinfo{journal}{\bibinfo{title}{A framework for the automatic
  generation of surface topologies for abdominal aortic aneurysm models}}.
\newblock {\emph{\JournalTitle{Annals of biomedical engineering}}}
  \textbf{\bibinfo{volume}{39}}, \bibinfo{pages}{249--259}
  (\bibinfo{year}{2011}).

\bibitem{antiga2008image}
\bibinfo{author}{Antiga, L.} \emph{et~al.}
\newblock \bibinfo{journal}{\bibinfo{title}{An image-based modeling framework
  for patient-specific computational hemodynamics}}.
\newblock {\emph{\JournalTitle{Medical \& biological engineering \&
  computing}}} \textbf{\bibinfo{volume}{46}}, \bibinfo{pages}{1097--1112}
  (\bibinfo{year}{2008}).

\bibitem{bertels2019optimizing}
\bibinfo{author}{Bertels, J.} \emph{et~al.}
\newblock \bibinfo{title}{Optimizing the dice score and jaccard index for
  medical image segmentation: Theory and practice}.
\newblock In \emph{\bibinfo{booktitle}{International Conference on Medical
  Image Computing and Computer-Assisted Intervention}},
  \bibinfo{pages}{92--100} (\bibinfo{organization}{Springer},
  \bibinfo{year}{2019}).

\bibitem{taha2015metrics}
\bibinfo{author}{Taha, A.~A.} \& \bibinfo{author}{Hanbury, A.}
\newblock \bibinfo{journal}{\bibinfo{title}{Metrics for evaluating 3d medical
  image segmentation: analysis, selection, and tool}}.
\newblock {\emph{\JournalTitle{BMC medical imaging}}}
  \textbf{\bibinfo{volume}{15}}, \bibinfo{pages}{1--28} (\bibinfo{year}{2015}).

\bibitem{derrick2016welch}
\bibinfo{author}{Derrick, B.}, \bibinfo{author}{Toher, D.} \&
  \bibinfo{author}{White, P.}
\newblock \bibinfo{journal}{\bibinfo{title}{Why welch’s test is type i error
  robust}}.
\newblock {\emph{\JournalTitle{The Quantitative Methods in Psychology}}}
  \textbf{\bibinfo{volume}{12}} (\bibinfo{year}{2016}).

\bibitem{yushkevich2006user}
\bibinfo{author}{Yushkevich, P.~A.} \emph{et~al.}
\newblock \bibinfo{journal}{\bibinfo{title}{User-guided 3d active contour
  segmentation of anatomical structures: significantly improved efficiency and
  reliability}}.
\newblock {\emph{\JournalTitle{Neuroimage}}} \textbf{\bibinfo{volume}{31}},
  \bibinfo{pages}{1116--1128} (\bibinfo{year}{2006}).

\bibitem{VTK4}
\bibinfo{author}{Schroeder, W.}, \bibinfo{author}{Martin, K.} \&
  \bibinfo{author}{Lorensen, B.}
\newblock \emph{\bibinfo{title}{{The Visualization Toolkit--An Object-Oriented
  Approach To 3D Graphics}}} (\bibinfo{publisher}{Kitware, Inc.},
  \bibinfo{year}{2006}), \bibinfo{edition}{fourth} edn.

\bibitem{meshlab}
\bibinfo{author}{Cignoni, P.} \emph{et~al.}
\newblock \bibinfo{title}{{MeshLab: an Open-Source Mesh Processing Tool}}.
\newblock In \bibinfo{editor}{Scarano, V.}, \bibinfo{editor}{Chiara, R.~D.} \&
  \bibinfo{editor}{Erra, U.} (eds.) \emph{\bibinfo{booktitle}{Eurographics
  Italian Chapter Conference}},
  \url{10.2312/LocalChapterEvents/ItalChap/ItalianChapConf2008/129-136}
  (\bibinfo{publisher}{The Eurographics Association}, \bibinfo{year}{2008}).

\end{thebibliography}

\section*{Acknowledgements}

The authors would like to thank Jane Liggins and Miriam Hayward, Intra Imaging for assisting with the collection of this data. SB would like to acknowledge the Auckland Academic Health Alliance (AAHA) and the Auckland Medical Research Foundation (AMRF) their financial support and endorsement. This research was undertaken with the assistance of resources from the National Computational Infrastructure (NCI Australia), an NCRIS enabled capability supported by the Australian Government. 

This research was conducted with approval from the University of New South Wales Human Research Ethics Committee (Ref. HC190145) and University of Auckland Human Participants Ethics Committee (Ref. 022961).

\section*{Author contributions statement}
RG, DA and KE have contributed to annotation of the ground truth data and collation of the dataset in consultation with AS and SO. RG has analysed the results. MW, CE and SB have established the Coronary Atlas which provided the data for this study. All authors reviewed the manuscript.

\section*{Competing interests}

The authors declare that they have no known competing financial interests or personal relationships which have or could be perceived to have influenced the work reported in this article

\section*{Figures \& Tables}

\end{document}